# An Energy-Efficient High Definition Map Data Distribution Mechanism for Autonomous Driving

Jinliang Xie, Jie Tang, *Member, IEEE* and Shaoshan Liu, *Senior Member, IEEE*

**Abstract** Autonomous Driving is now the promising future of transportation. As one basis for autonomous driving, High Definition Map (HD map) provides high-precision descriptions of the environment, therefore it enables more accurate perception and localization while improving the efficiency of path planning. However, an extremely large amount of map data needs to be transmitted during driving, thus posing great challenge for real-time and safety requirements for autonomous driving. To this end, we first demonstrate how the existing data distribution mechanism can support HD map services. Furthermore, considering the constraints of vehicle power, vehicle speed, base station bandwidth, etc., we propose a HD map data distribution mechanism on top of Vehicle-to-Infrastructure (V2I) data transmission. By this mechanism, the map provision task is allocated to the selected RSU nodes and transmits proportionate HD map data cooperatively. Their works on map data loading aims to provide in-time HD map data service with optimized in-vehicle energy consumption. Finally, we model the selection of RSU nodes into a partial knapsack problem and propose a greedy strategy-based data transmission algorithm. Experimental results confirm that within limited energy consumption, the proposed mechanism can ensure HD map data service by coordinating multiple RSUs with the shortest data transmission time.

## I. INTRODUCTION

AUTONOMOUS driving is one of the most promising applications in artificial intelligence and intelligent transportation. It liberates human drivers, alleviates traffic congestion, offers great convenience as well as driving safety, and finally reduces environmental pollution [1]. High Definition Map (HD map) is one basis for autonomous driving [2], [3]. Compared with traditional electronic maps, HD map has the characteristics of high-precision data, high-dimensional information, and high real-time response. With the extension of sensing range and the improvement of sensing accuracy, HD map can build up a more comprehensive autonomous driving guidance in the way of perception, positioning, and decision-making [4], [5].

By HD Map service, we can have the centimeter-level high-precision positioning for autonomous driving vehicles [6]. Using prior knowledge in map, the HD Map can reduce the difficulty of environment perception [7]. It can also broadcast various road information in advance, realize more effective driving planning like early vehicle deceleration and obstacle avoidance. Besides, HD map supports personalized driving, including various driving behavior suggestions, the best acceleration point suggestions, the best cornering speed suggestions, etc. All of them can be used to improve the safety and comfortability of autonomous driving.

In this paper, we present a mechanism that provides real-time HD MAP services for autonomous driving under tight real time and energy budget. While literature is rich with data provision designs under Internet of Vehicle networking, prior efforts are mostly targeting ordinary command data in small volume. For example, early warning data, traffic notification and instruction data. However, we find that the HD Map service for autonomous driving operates under totally different scenarios. It works with great constraints in high responsibility and high data volume. For instance, as the supplement of real-time perception, the response of HD map should be less at least than 80ms on average to fit in autonomous driving pipeline [8] and its data volume per second can be over 2G per second in some high speed scenarios.

With this background, we **firstly quantitatively demonstrate that the existing data distribution services does not fit the scenarios for HD map service** (Sec. 3). The in-vehicle disk+memory solution can support the HD map service for only few minutes. Therefore, maps are usually being served by online services, and only a nearby small areas of the HD Map (called submaps) are downloaded to the vehicle when needed. In real-time V2V data transmission,

Xie, J., is with the South China University of Technology, Guangzhou 510641, China. (e-mail: 201821033740@mail.scut.edu.cn).
Tang, J., is with the South China University of Technology, Guangzhou 510641, China. (e-mail: cstangjie@scut.edu.cn). She is the corresponding author of this paper.
Liu, S., is with PerceptIn, Santa Clara, CA 95054. (e-mail: shaoshanliu@perceptin.io).

theoretically there need to contact up to hundreds of cars to fill the map data tank and thus posing a great challenge for limited peer-cars and limited bandwidth in real-world scenarios. In V2I based solution, with proactive data provision, it can deliver the feasible HD Map service. However, it leaves a question that how the service can be provided with multiple RSU nodes in cooperation.

With data transmission scenarios carefully calculated, we propose **a HD map data distribution mechanism on top of V2I data** transmission (Sec 4.). The proposed solution considers the energy consumption of vehicles as well as the bandwidth sharing between them. By this mechanism, the map provision task is allocated to the selected RSU and sector of HD map data is transmitted proportionately.

We model the selection of RSU nodes into **a NP-hard back-pack problem and divide the multi-agent question into multiple one-agent question** (Sec. 4). We further propose a heuristic algorithm to solve the selection by the complexity in *O(logn),* which ensures that the decision of node selection and their data task allocation can be output in real-time.

To show its generality, we discuss how RSUs density, traffic condition, data size can impact the efficiency of the proposed mechanism (Sec.5). Experimental results show that the mechanism performs better in both time and energy. Meanwhile, the number of RSU nodes can be greatly reduced, when there is much traffic, the RSU nodes hit is distributed more evenly.

The main contributions of this paper are:
1) We quantitatively show that no single mechanism fits the demands and constraints of HD Map service. The analysis provides a desirable acceleration target for the HD map service.
2) We propose a V2I based HD map distribution mechanism. It assigns map data provision task to selected RSU nodes and makes them co-complete the map transmission for target area.
3) We model the data provision task designation into a partial backpack problem and design a heuristic algorithm to solve it in the complexity of *O(logn)*.
4) We run the simulation experiments and verify the performance of proposed mechanism

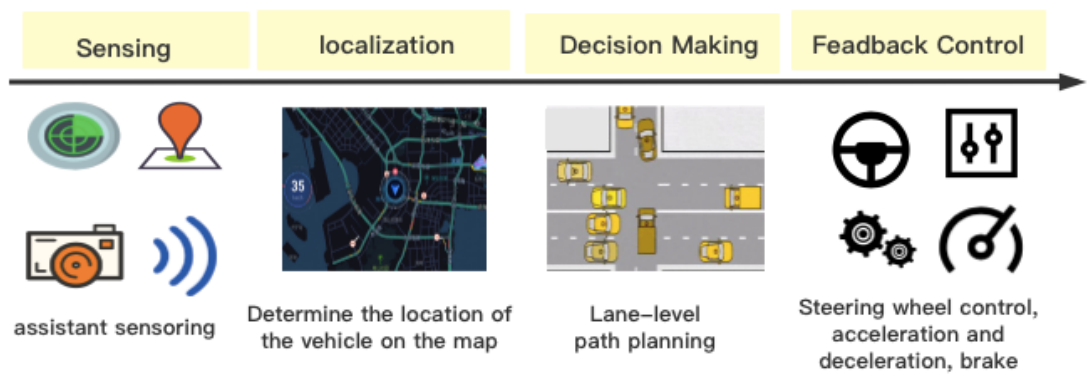

Fig. 1. The role of HD map in autonomous driving

TABLE I
The HD Map Information

| The map content | Purpose | Data Source | Data size (per mile) |
|---|---|---|---|
| *2D Orthographic Reflectivity Map* | *Localization* | *Lidar* | *>5 GB* |
| *Digital Elevation Model (DEM)* | *Planning&Sensing* | *Sonar/Radar* | *100 MB* |
| *Lane/Road Model* | *Planning&Localization* | *Camera/ Sonar* | *10 MB* |
| *Semantic Layer* | *Planning&Sensing* | *Camera* | *50 MB* |

II. BACKGROUND OF HD MAP

HD maps are the foundation for autonomous driving implementation. It usually has multiple layers and provides a full stack of information for autonomous vehicles. Layers of HD Maps have different representations, data structures and purposes. Usually, HD Maps contain the following 4 layers [9] and we show each size in Table I:

1) **2D Orthographic Reflectivity Map**

This layer is a 2D planar view of the road surface extracted from the LiDAR 3D point clouds [10], [11].

2) **Digital Elevation Model (DEM)**

DEM is a 3D model and contains the height information of surface of the driving environment, such as the height of the road curbs, the grade/steepness of a ramp or hilly road etc [12].

3) **Lane/Road Model**

Lane/road model contains the semantics of lane segments and road segments. Lane model contains information of lane geometrics (boundaries, width, curvature etc.), lane type (car lane, bike lane, bus-only lane etc.), lane directionality, lane marking/divider types (solid vs dashed, single vs double etc.), restrictions (e.g. left/right turn only), speed limits, connectivity between lanes, etc. Lane/road model is critical for motion planning, vehicle control etc.

4) **Semantic Map**

It is usually a versatile layer that stores the semantics of static elements in the driving environment, e.g. traffic lights and their association with lanes, road obstacles etc.

## III. MOTIVATIONS

### A. Overview on Data Distribution Solutions

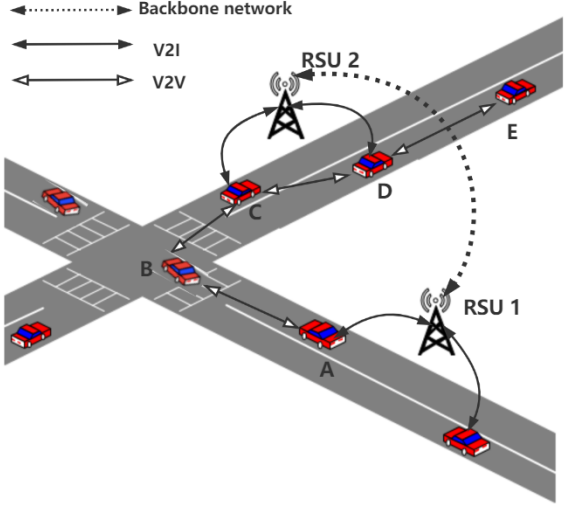

Fig. 2. V2V and RSU cooperative transmission (If vehicle D would like to get data from A: by V2V, the data will be sent by multi-hop through A->B->C->D; by V2I, A->RSU1-> RSU2 ->D )

The data distribution methods in vehicle application can be roughly divided into three types: V2V (Vehicle to Vehicle), V2I (Vehicle to Infrastructure), and V2X (Vehicle to Vehicle or Infrastructure), that is a fusion of V2V and V2I.

V2I is the communication between the vehicle and the infrastructure by roadside, like RSU.

V2V is used for the communication between cars. The DSRC based method is the mainstream in the data distribution by V2V now. V2V uses the mechanism of "carry store forward". The vehicle can receive data from the nearby vehicles, carry the data in its traveling and make another data forwarding when it meets some target vehicle.

In most cases, V2I will cooperate with V2V to form a V2X data distribution solution. RSU can be connected to one or more Traffic Information Centers (TIC) through the backbone network. It greatly increases the amount of data transmitted in the vehicle network. V2V increase the chance that data can be successfully delivered to dedicated destination.

### B. In-Vehicle Storage for HD Maps

Autonomous vehicles consume HD Map data in a considerably high rate. It can be seen from Table 1 that the amount of map data consumed per second during driving is over 300MB. However, that is the data cost from only one of optional roads in planning. In driving, the vehicle will be fed with the map data for all possible planning roads in some destination range. Thus, from a storage point of view, such consumption will digest 4T hard disk data quickly in just half hour. Therefore, it is impossible for the vehicle to store all the map data of the target area in advance. HD map data should be divided into submaps and provided by some online service.

From the perspective of temporary data caching, the downloaded submaps will be erased right after the cars drive out the area those data maps to. Due to the limited caching room, the used map data will become invalid in a very short time. As a result, the original "carry-store-forward" data transmission method in V2V cannot work effectively.

### C. Shortage of v2v Data Transmission

1) The Analysis for ideal situation

In the V2V transmission, it is assumed that there are two vehicles $Car1$ and $Car2$ running at a constant speed $v1$ and $v2$ in a bidirectional lane of a highway. Here, $Car1$ is the source vehicle, $Car2$ is the target vehicle. If the vehicle uses DSRC communication and its communication range is $r$ (generally 100-200 meters), the distance between the two vehicles in their driving direction is $d$ ($d<r$), the vehicle data transmission rate is $R$.

According to 3-1, the contact time for them within their communication range is $T$, and the amount of data transmitted during this time is $C$:

$$T = \frac{2\sqrt{r^2 - d^2}}{v_1 + v_2} \quad (3-1)$$

$$C = TR \quad (3-2)$$

Here, $R$ is a function of the vehicle distance $D$, the DSRC channel bandwidth $B$, the DSRC channel attenuation coefficient $h$, the channel noise power spectral density $N_0$ and the power for vehicle data transmission $P_S$. To simplify the model, it assumes $R_{max} = B$, we can get $C_{max} \leq \frac{2\sqrt{r^2-d^2}}{v_1+v_2} \times R_{max}$. Then we have:

$$R = Blog_2\left(1 + \frac{DP_S|h|}{N_0}\right) \quad (3-3)$$

When the amount of map data to be transmitted is $G$, in an ideal situation, there is always a vehicle participating in data transmission and the carry-store-forward of data is always working, we need to have $Num = G/C_{max}$ vehicles to travel at lease $Num \times VT$ mile for map data acquisition.

2) The Analysis for Actual situation

In the real-life situation, if the data is received from vehicles in the opposite direction, the probability that vehicles participate in the transmission is $p = \frac{mr}{vt}$. When there has $m$ vehicles in the reverse lane within time duration $t$. Thus, the distance used to complete the data transmission is $\frac{Num \times VT}{P}$.

If the V2V data is obtained from cars met at different lanes in the same direction, where they travel at varies speeds. Only in this way, it can have can different data fragments from different vehicles. The contact time of the two vehicles becomes $T = \frac{2r}{|v_1 - v_2|}$, then there requires a distance of $\frac{Num \times VT}{P}$ to transmit $G$ data.

### D. V2I

Compared with V2I, the data transmission coverage of infrastructure node is much wide. Normally, it is 3.X-5.X

wider compared with the one of V2V. Thus, the contact time of them can be extended correspondingly. However, there still needs couples of RSU to participate in the map transmission due to its large volume. RSU nodes are always connected by a backbone network, so that one can get a data copy from other nodes within a trivial time slice. Therefore, the problem in V2I turns to be how some data distribution mechanism can reasonably designate transmission tasks to those RSUs.

*E. Conclusion and Observations*

From the above analysis, we can get some observations:

First, in the case of limited in-vehicle storage, vehicles must use online data provision solutions.

Second, the amount of map data is large, it is impractical for the vehicle to obtain data only through V2V. V2V's high energy consumption also cannot support this degree of data transmission.

Third, the downloaded map data will be invalid in a very short time, which makes the original "carry-store-forward" V2V transmission impractical.

Last, V2I is a comparative feasible solution for HD map data service, however, an effective and efficient data distribution mechanism is required, and this is the main purpose of this paper.

## IV. MAP DATA DISTRIBUTION MECHANISM

*A. V2I Based Solution*

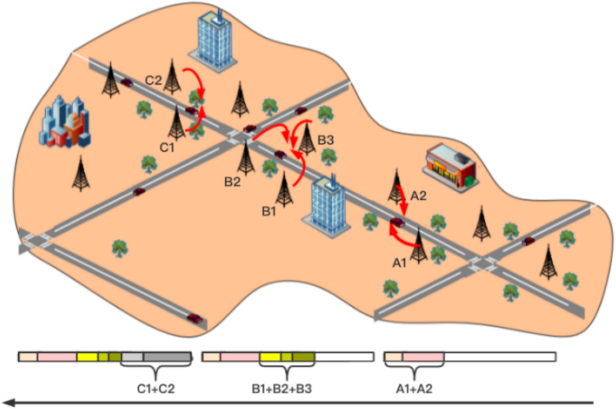

Fig. 3. The overview for V2I based HD Map Service

V2I based map data transmission exhibits great potentials in serving HD map data in real time. However, there may co-exist several vehicles which all require the map data from the same RSU node concurrently. When the number of vehicles exceeds some threshold, the data transmission will be much slowed down due to the transmission rate inference. Therefore, the vehicle must continue to search the next possible data source.

Thus, as stated in Fig 3, the V2I based solution should be able to find a series of $\alpha_i$, $0<\alpha_i<1$ & $\sum \alpha_i =1$ for each RSU node in the area, which is responsible for providing map data service proportionally to the designated $\alpha_i$. With well consideration of constraints like resource sharing and energy cost, it aims to have the transmission time minimized without breaking the energy budget.

*B. Modeling the Multi-vehicle map data distribution*

Supposing there are in total $m$ car in the lane that require HD map data, the amount of data that each vehicle needs to transmit is $M$, and the remaining energy of the vehicle $j$ is $w_j^{car}$. Assume that all vehicles are electronical, the energy consumption of a traveling electric vehicles is [13]:

$$E = \frac{\int_0^{t_0} P_e dt}{S} \qquad (4-1)$$

Where $E$ is the energy consumption per unit mileage, $t_0$ is the driving time, $S$ is the driving mileage and $P_e$ is the power requirement of driving a car.

Energy consumed by vehicle $j$ receiving map data from RSU node $i$ is $E_{i,j}^g$. As shown below, it is a function of the amount of data, the vehicle transmission rate, and the distance between RSU node and vehicle lanes.

$$E_{i,j}^g = f(M_i, R_{i,j}^g) \qquad (4-2)$$

From the above analysis, the cost that vehicle can autonomously navigate through the area is the sum of the energy used for driving through it and the energy for the map data transmission of the area. Assuming that the remaining energy of car $i$ is $W_i^{car}$, to enable autonomous driving, it needs to ensure that:

$$\sum E_{i,j}^g + \int_0^{L_j} E \, ds < W_i^{car} \qquad (4-3)$$

The transmission rate is related to the bandwidth, transmission power, transmission distance, and path loss. According to the literature [14], it is assumed that the maximum downlink transmission power of the edge node is $P_{max}^B$, and $P_{i,j}^B P_{i,j}^B$ represents the downlink transmission power allocated by the edge node $i$ for its communication with vehicle $j$, here $0 \leq P_{i,\ j}^B \leq P_{max}^B$. Path fading can be expressed as $d_i^{-\sigma}$, where $d_{i,j}$ and $\sigma$ respectively represent the distance between RSU node $i$ and vehicle $j$, as well as the path loss coefficient. The receiving channel bandwidth of vehicle $j$ is $B_j^g$, the attenuation coefficient of the receiving channel is $h_1$, and the power spectral density of noise in the channel is $N_0$. According to Shannon's formula[15], the rate at which vehicle $j$ uses to receive data from edge node $i$ is:

$$R_{i,j}^g = B_j^g log_2 \left(1 + \frac{d_i^{-\sigma} P_{i,\ j}^B |h_1|}{N_0}\right) \qquad (4-4)$$

Under the multi-vehicle model, it is assumed that the probability of any two vehicles meeting within the coverage of the same RSU node is $p$. Thus, the probability of $k$ vehicles simultaneously in the same RSU follows the Poisson distribution:

$$P(X = k) = \frac{e^{-mp}(mp)^k}{k!} \qquad (4-5)$$

Since multiple vehicles will compete for the bandwidth, they will be interfered by the surrounding vehicles when transmitting data. Thus, the transmission rate will be:

$$R_{i,j}^G(X = k) = B_j^g log_2 \left(1 + \frac{d_i^{-\sigma} P_{i,\ j}^B |h_1|}{N_0 + \sum_{j' \neq j}^{k} d_i^{-\sigma} P_{i,\ j'}^B |h_1|}\right)(4-6)$$

According to the analysis, the multi-vehicle map service problem with energy limits can be transformed into the following optimization problem. That is to minimize total data transmission time with the subject to the energy limitation.:

$$min \ max\left\{\sum_{i=1}^{N} t_i^j, j = 1,2,...,m\right\} \quad (4-7)$$

subject to:

$$\sum_{i=1}^{N} t_i^j \left(\sum_{k=1}^{m-1} P(X=k) R_{i,j}^g(X=k)\right) \geq M_j \quad (4-8)$$

$$0 \leq t_i^j \leq T_i, i = 1,2,...,N \quad (4-9)$$

$$\sum_{j=1}^{m} E_{i,j}^g < W_j^{car} \quad (4-10)$$

The final result is the tuple $<c_i^j, t_i^j>$ for each vehicle. $c_i^j$ is an expression in <0,1>, representing whether vehicle $j$ will transmit data transmission with RSU node $i$ when it passes by. $t_i^j$ represents the time duration vehicle $j$ used to transmit data with node $i$.

### C. Multi-vehicle Map data distribution solution

**To solve this model, we can first decompose the multi-vehicle problem into a single-vehicle problem.** The, we can find the optimal solution locally, and combine all the solutions together to get a globally optimal solution for the multi-car problem. For the solution of the single-vehicle problem, it can be assumed that there is only one vehicle $j$ obtaining data from *<RSU List>*. Each RSU node in *<RSU List>* can be selected or not be selected, and the total data amount transmitted in its limited time is no less than M. Besides, the remaining energy consumption of the vehicle must ensure driving safety. Thus, it can be considered as a variant of a knapsack problem.

The backpack problem can be divided into 0-1 backpack problems and partial backpack (fractional knapsack) problem. In our transmission model, even if the RSU node is selected, only part of the map data can be transmitted, so our question on map distribution can be regarded as a partial backpack problem. The partial backpack problem itself is an NP-complete problem with a complexity of ***O(2ⁿ)***, so it cannot find out an optimal solution by force.

We propose a heuristic algorithm named ETDM (Energy-Time Distribution Mechanism) that uses the greedy strategy to find the optimal solution. The idea of the ETDM is to sort the optional RSU nodes in descending order by the average transmission rate each gives. The optimal complexity of the sorting algorithm can be ***O(nlogn)***. After sorting, a new bandwidth sequence is obtained ($R'_1, R'_2, ..., R'_k$), then the transmission tasks can be allocated proportionally according to the bandwidth from high to low. The RSU node finally assigned some the transmission task may not use up its time contacting with vehicles.

To this end, we should find a time-optimal solution for a fractional backpack problem and combine them into a solution for multi-car model. The maximum value of the optimal solution of all vehicles will be taken as a result for the model of multi-vehicles. Because we should find the optimal time for m vehicles, thus the algorithm complexity is ***O(mnlogn)***. The pseudo-code of the algorithm is as follows.

---
**Algorithm 1** Multi-Vehicle data distribution based on energy consumption
---
**Input:** $B_1^g, B_2^g, ..., B_k^g$,
  $d_1^{-\sigma}, d_2^{-\sigma}, ..., d_k^{-\sigma}$,
  $P_{1,j}^B, P_{2,j}^B, ..., P_{k,j}^B$,
  $h_1, N_0, W_i^{Car}, L_j, E, m, M_{basic}$
**Output:** $t$
1: **if** $\int_0^{L_j} Eds > W_i^{Car}$ **then**
2:   $M_j \leftarrow M_{basic}$
3:   $P(X = x) \leftarrow \dfrac{e^{-mp}(mp)^k}{k!}$
4: **end if**
5: **for** $i = 1 \to k$ **do**
6:   $R_{i,j}^G(X = x) \leftarrow B_j^g log_2(1 + \dfrac{d_i^{-\sigma} P_{1,j}^B |h_1|}{N_0 + \sum_{i' \neq i}^{x} d_{i'}^{-\sigma} P_{i',j}^B |h_1|})$
7:   $R_{i,j}^g \leftarrow \sum_{x=1}^{m-1} P(X = x) R_{i,j}^G(X = x)$
8: **end for**
9: $E_{i,j}^g \leftarrow f(M_i, R_{i,j}^g)$
10: $E_{total} \leftarrow \sum E_{i,j}^g + \int_0^{L_j} Eds$
11: **if** $E_{total} > W_i^{Car}$ **then**
12:   $M_j \leftarrow M_{basic}$
13: **end if**
14: **for** $j = 1 \to m$ **do**
15:   $R' \leftarrow DescSort(R_{1,j}^g, R_{2,j}^g, ..., R_{k,j}^g)$
16:   $M'_1 \leftarrow R_{1,j}'^g T_1, i \leftarrow 1$
17:   **while** $M'_i < M_j$ **do**
18:     $M'_{i+1} \leftarrow M'_i + R_{i+1,j}'^g T_{i+1}$
19:     $i \leftarrow i + 1$
20:   **end while**
21:   $t_i^j \leftarrow \dfrac{M - \sum_{k=1}^{i-1} R_{k,j}'^g T_k}{B'_i}$
22:   **while** $--i > 0$ **do**
23:     $t_i^j \leftarrow T_i$
24:   **end while**
25:   Other $t^j \leftarrow 0$
26: **end for**
27: $t \leftarrow max\{\sum_{i=1}^{k} t_i^j\}, j = 1 \to m$

---

## V. EXPERIMENT AND CONCLUSION

### A. Experiment setup

In this paper, we implement the simulation on MATLAB platform, and the traffic data is extracted from a 10-minute traffic segment of a crossroad from the **Luxembourg SUMO** traffic (LuST) data set [16]. The vehicles drive through the crossroad into three individual areas, areas A, B, and C. A total of 251 vehicles were detected within 10 minutes, where there had 95 vehicles going into area A, 94 vehicles going into area B, and 62 vehicles going into area C. Here, we assume the vehicle power of data transmission is 10W and the vehicle battery consumption rate is between 0.15kwh and 0.25kwh. We randomly generate 60 RSU nodes, and all parameters of each RSU node like bandwidth, coverage radius, path attenuation coefficient in transmission, RSU transmission channel noise, etc. were all randomly generated within their reasonable range. For example, the bandwidth is set to be 1 ~ 2GBps, the coverage radius is 100 meters, the average distance between each RSU node is 100 meters, the distance between the RSU node and the road lane is 20 ~ 70 meters, and the RSU transmission channel noise power spectral density is set to be 0.3.

### B. Data distribution experiment

The baseline of the evaluation includes: The Original Algorithm (OA), which is to transmit the map data by the sequence that vehicles arrives the RSU nodes. Every time a vehicle enters into the transmission range an RSU node covers, as long as it contains the required map sectors, the data transmission can be started. In the Probabilistic Transmission Algorithm (PTA), as long as the RSU node contains the required map data, it starts to transfer data by some probability. For example, in a 0.7 probability transmission algorithm, the probability that the vehicle communicates with RSU node is 0.7. In our experiments, we configure the probability in 0.3/0.5/0.7 respectively.

The evaluation metrics include the optimal transmission time, the average transmission time of each vehicles under different traffic, the transmission time of a single vehicle under different traffic, the number of RSU nodes involved in each transmission, and the map data hit rate in RSU nodes, which is calculated by the variance of the hit rate of all nodes. We also evaluate how the amount of data and the location of the RSU nodes impact on the ETDM algorithm

#### 1) Influence of the amount of transmitted map data

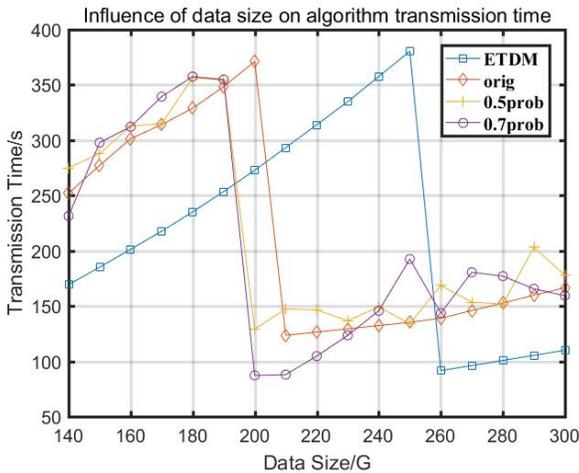

Fig. 4. Influence of data volume on algorithm transmission time

First, we evaluate the impact form the data volume. We have simulated 251 vehicles, and all of them have communicates with the same 60 RSU nodes. We gradually increase the amount of required data in each experiment, which increases from 140G to 300G, and the remaining energy capacity of the vehicle is set to be 5KWh.

From Fig.4, we can see when the data volume increases into above 190G, all algorithms have a noticeable cliff-like decline in transmitted data amount. That is because the energy capacity of the vehicle is not enough to support full-scale data transmission. At that time, the vehicle turns to transmit just the underlying layers of map, so there comes a declining cliff in data volume.

As stated in Fig. 5, when the vehicle transmits the same amount of data before the cliff, the proposed ETDM algorithm can significantly save transmission time by 34.8% on average. The number of RSU nodes the ETDM algorithm used is also on average 6-7 nodes less than the other three algorithms.

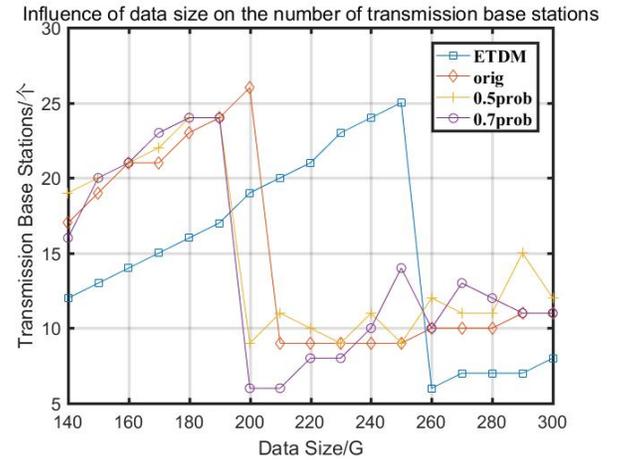

Fig. 5. Influence of data volume on the number of transmission base stations

#### 2) Influence of traffic flow on transmission time

The experiment on traffic is implemented by increasing the number of vehicles from 10 vehicles to 250 vehicles by adding 10 vehicles each time. We can find the shortest transmission time and the average transmission time for all vehicles in each configuration. Here, we only count the vehicles that have finished full map data transmission.

From the experimental results, no matter it is an optimal transmission time (Fig. 6) or an average transmission time(Fig. 7), the performance of the ETDM algorithm is superior to the other three: when taking about the optimal time, the ETDM algorithm saves time in 24.5%; when talking about the average transmission time, the ETDM algorithm offers 21.1% time saving compared with the OA and 18.2% time saving compared with the PTA. The Fig. 8 shows the effect of different traffic on the transmission time for a single vehicle. It can be seen that one vehicle can save an average of 24.1% transmission time if it applies the ETDM algorithm.

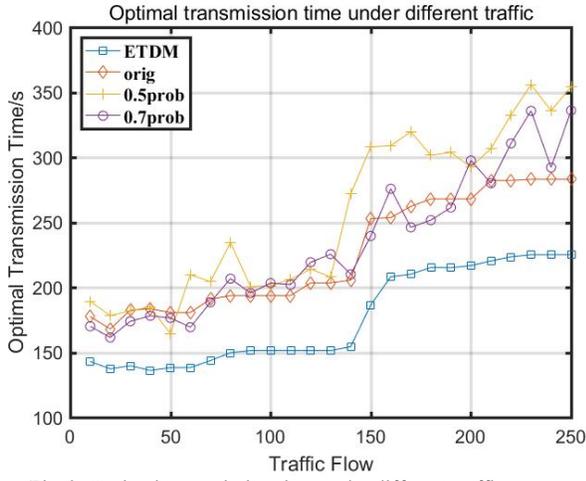
Fig.6. Optimal transmission time under different traffic

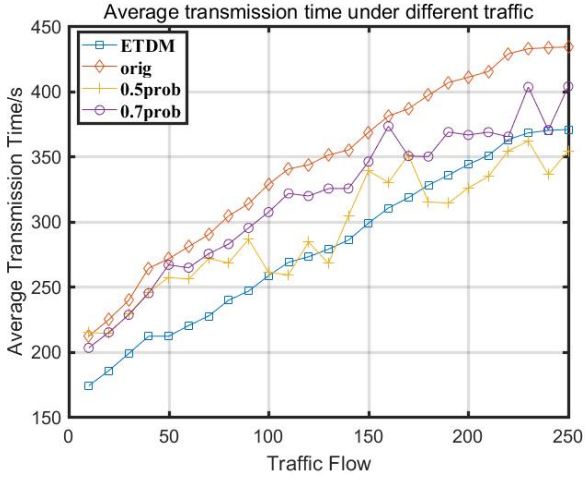
Fig. 7. Average transmission time under different traffic

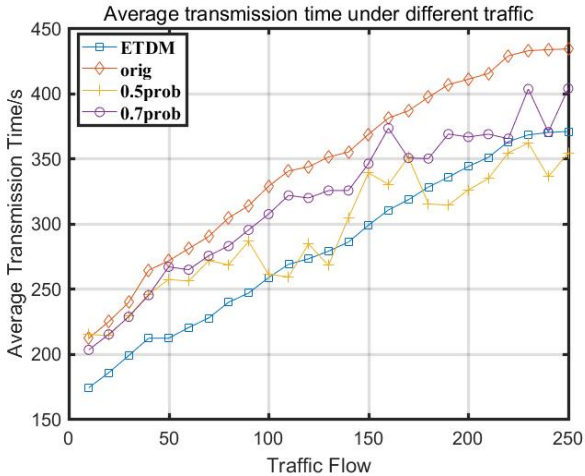
Fig.8. The transmission time of single car under different traffic

*3) Influence of traffic on usage of RSU node*

How many accesses an RSU node gets depends on the current traffic condition. In theory, the nodes with faster transmission speed will be accessed more frequently. Fig. 9 shows the statistics of the number that edge node is accessed under different traffic, as well as the variance of the hit rate in Fig. 10. It can be seen that when the traffic is light, the number of accessed nodes under the ETDM algorithms is much less than that of other three algorithms, but its variance of the hit rate is larger than others. This is because when the traffic is light, the ETDM tends to select the nodes having a larger bandwidth. As a comparison, the OA uses the FCFS thus its hit rate variance is prone to be large; and the PTA selects each node with the same probability thus its hit rate variance is comparatively small. However, when the traffic is heavy, the variance of the hit rate of the ETDM algorithm turns to be the smallest one, indicating that the ETDM algorithm has a relatively evenly distributed RSU hit rate

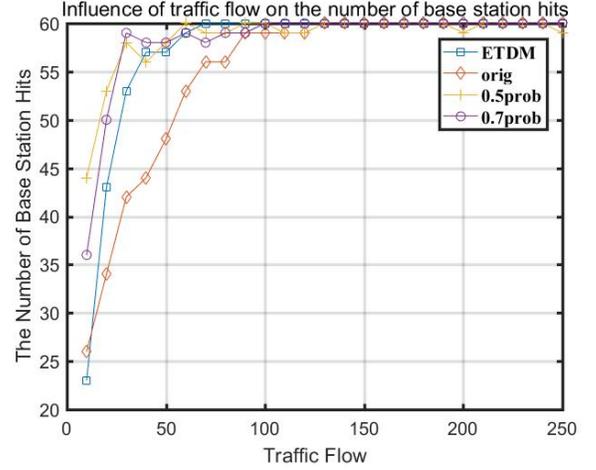
Fig. 9. Influence of traffic flow on the number of hit base station

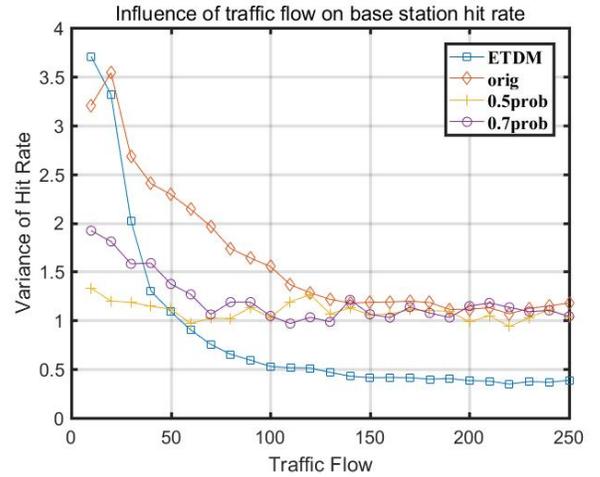
Fig. 10. Influence of traffic flow on base station hit rate

## VI. RELATED WORK

The previous data service in the Internet of Vehicles is mainly focusing on the data content distribution and content caching deployed by RSU. In [17], the author considered the position of the access points as well as the moving path of vehicles, then decided a way how data can be forwarded from a stationary point to a moving vehicle. It also comprehensively selected the locations for data transmission. In [18], authors use network coding techniques to give efficient data distribution between multiple RSUs. In [19], it studied the cost-effective planning problem of heterogeneous vehicle networks, that is composed of traditional macro base stations and RSUs with buffering supports. Literature [20] focused on the cloud based VANET architecture and its content retrieval. It proposed a cache placement strategy that considering the on-vehicle caching layer and RSU caching layer. In [21], the author introduced the problem of content distribution and caching in RSU, as well as proposed a greedy algorithm to traverse the cached data with reasonable cost. Literature [22] studied the deployment of RSU in the highway. It comprehensively considered factors of wireless interference, traffic distribution, and vehicle speed to maximize the data throughput in IOV. Similar to literature [22], [23] studies the deployment of RSU in urban scenarios. The author comprehensively modeled the traffic in the urban as well as the probability of accidents at different locations. By designing a position awareness mechanism, such data service can efficiently minimize the accident reporting time in the city.

There is little research on content distribution and caching in V2V communication. For data caching, they mainly talked about its application in Content Centric Networks (CCN) or Named Data Networking (NDN). In [24], author proposed a node selection algorithm based on the theory of minimum vertex cover set in a static network and proposed a cooperative cache mechanism. It used social relations to select cache points and solve the connection failure caused by vehicle movement. Literature [25]-[27]introduced many basic caching strategies in CCN, but most of these caching strategies are more suitable for static network scenarios. Literature [28]-[30] studied the use of caching technology in wireless networks to support user mobility and improve content distribution. Literature [31] reviewed the opportunities and challenges of edge computing for autonomous driving from a security perspective, reviewed state-of-the-art approaches in these areas, and discussed potential solutions to these challenges.

VII. CONCLUSION.

HD map is the basis for autonomous driving and map data service is crucial for real-time and safety for autonomous driving. In this paper, we quantitatively identified that the existing data transmission solutions cannot guarantee the quality of HD map data service. To solve this problem, we propose a HD map data distribution mechanism on top of V2I data transmission. Our proposed method divides the map provision task to selected RSU nodes and has map data proportionately transmitted. By multiple RSUs cooperation, the HD map service can be delivered in-time with optimized in-vehicle energy consumption. Here, the problem that map data distributed between nodes is modeled into a partial knapsack problem and we propose a greedy algorithm ETDM as solution. ETDM turns the multi-agent question into a multiple one-agent question and can generate the allocation decisions with complexity of *O(logn)*. Experimental results show when the remaining energy capacity of the vehicle is limited, the proposed method can provide shorter map data transmission time, more balanced RSU accesses even when the traffic is heavy.